\documentclass[10pt,twocolumn,letterpaper]{article}

\usepackage{cvpr}              

\usepackage{graphicx}
\usepackage{booktabs}
\usepackage{adjustbox}
\usepackage{tabularx}
\usepackage{bbm}
\usepackage{xcolor}
\usepackage{wrapfig}
\usepackage{amssymb}
\usepackage{amsmath}
\usepackage{multirow}
\usepackage{array}
\usepackage{soul}

\definecolor{cvprblue}{rgb}{0.21,0.49,0.74}
\usepackage[pagebackref,breaklinks,colorlinks,allcolors=cvprblue]{hyperref}

\def\paperID{10381} 
\def\confName{CVPR}
\def\confYear{2026}

\usepackage[table]{xcolor}
\definecolor{lightgreen}{HTML}{CCFFCC} 

\newcommand{\OURS}{\text{SCAPO}\xspace}

\title{SCAPO: Self-Supervised Category-Level Articulated Pose Estimation from a Single 3D Observation}

\author{Can Zhang \qquad Gim Hee Lee\\
Department of Computer Science, National University of Singapore\\
{\tt\small can.zhang@u.nus.edu \qquad gimhee.lee@nus.edu.sg}
}

\begin{document}
\maketitle
\begin{abstract}


Existing methods for category-level object articulation from a single 3D observation often rely on dense supervision, multi-frame inputs, or CAD templates, and still struggle to disentangle geometry from articulation or to recover explicit joint parameters. We propose \OURS, a self-supervised framework that estimates canonical geometry, rigid part segmentation, and joint pivots, axes, and articulation states from a single RGB-D observation without ground-truth labels or category-specific models. Our \OURS first uses an $\operatorname{SE}(3)$-equivariant vector-neuron autoencoder to factor out global pose and align diverse instances into a shared canonical space. On this aligned shape, a joint-aware blend-skinning module is then designed to model part motion. We learn this representation through cycle reconstruction between observed and canonical shapes and cross-space alignment with a learnable canonical template that decouples shared category geometry from instance-specific residual shape. Experiments on synthetic and real articulated-object datasets show that our \OURS recovers consistent part structure and accurate articulation parameters and outperforms all self-supervised baselines. 
Our source code is available at: \url{https://lulusindazc.github.io/SCAPOproject/}.
\vspace{-8mm}

\end{abstract}    
\section{Introduction}
\label{sec:intro}

Articulated objects consist of multiple rigid parts connected by joints such as doors, drawers and laptop screens. They appear in most everyday environments and often govern how humans act on the world. For robots and embodied agents, structured understanding of these objects requires part-level pose, segmentation and joint parameters such as rotation axes or sliding directions. This task is especially challenging for previously unseen instances within a known category and becomes even harder when ground-truth annotations or predefined 3D templates are absent. In this paper, we study the problem of estimating part segmentation, per-part pose and joint properties from a single RGB-D image of an articulated object under a self-supervised setting. Success in this setting would benefit robotic manipulation, AR/VR and digital-twin systems, where reliable kinematic reasoning from sparse observations is crucial.

\begin{figure}[t]
\centering
\includegraphics[scale=0.28]{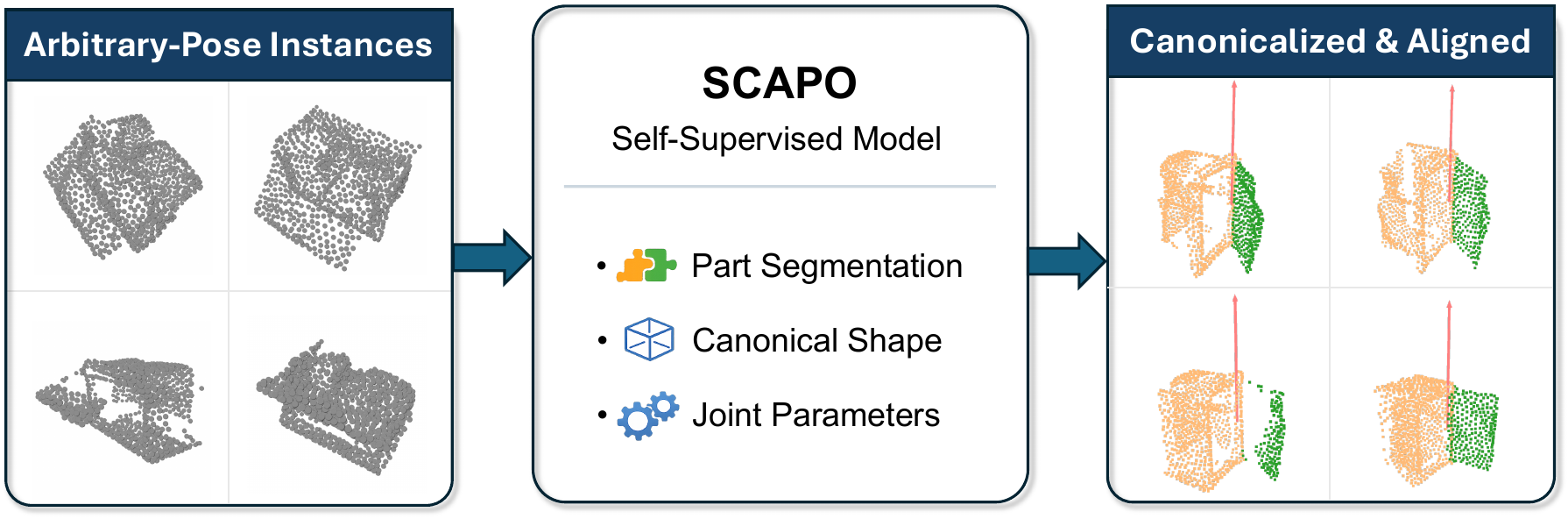}
\caption{Our model maps articulated objects under arbitrary poses to global poses, canonical shapes, part segmentation and joint parameters in a self-supervised framework.} \vspace{-4mm}
\label{fig:teaser}
\end{figure}

Early work on articulated modeling~\cite{hausman2015active,martin2016integrated} often relies on instance-level supervision with CAD models and joint annotations at test time, which limits generalization to new instances. Category-level articulated object modeling removes this restriction and aims to predict part poses and joint attributes for novel objects within a category under a shared kinematic structure. 
A self-supervised formulation is especially attractive because it avoids expensive labels for part poses, segmentations and joint metadata. However, it must cope with large intra-class shape variation, ambiguous part boundaries, unknown articulation states and kinematics, and strong ambiguity from single-frame observations.

Recent works tackle self-supervised category-level articulation from several angles. PartMobility~\cite{shi2021PartMobility} uses motion cues in point cloud sequences to discover moving parts and infer motion attributes. However, it depends on dynamic input and does not recover full object or part poses, which limits deployment in static scenes. UPPD~\cite{kawana2022uppd} handles static inputs by decomposing shapes into parts with $\operatorname{SE}(3)$ transforms and implicit fields. Nonetheless, its reliance on volumetric supervision weakens robustness on sparse or noisy point clouds and makes articulation hard to separate from shape variation. EAP~\cite{liu2023EAP} learns part-level $\operatorname{SE}(3)$-equivariant features to disentangle pose and shape and generalizes better to unseen instances. However, it only predicts latent part transforms and lacks explicit joint pivots or motion axes, which restricts downstream kinematic reasoning. 
OP-Align~\cite{che2024opalign} introduces a self-supervised framework that aligns object poses and part articulations via canonical reconstruction from a single point cloud, and shows strong category-level generalization. 
Nonetheless, it still struggles with fine-grained articulation and large shape variation because geometry and articulation remain entangled. Despite the progress, existing methods still face key gaps in recovering explicit joint structure from a single observation, and in disentangling geometry and motion across diverse real-world objects.

We propose \OURS, a self-supervised framework that predicts canonical shape, part segmentation and joint parameters from a single 3D observation of an articulated object. Our \OURS first learns a category-level canonical space with an $\operatorname{SE}(3)$-equivariant autoencoder that factors out global pose and extracts pose-invariant shape features. These features align unseen instances into a shared canonical frame. Subsequently, our \OURS models part structure and motion with a joint-aware blend-skinning deformation that comprises a keypoint network to infer part centroids as bone anchors, a pose network to regress joint pivots, axes and articulation states, and a skinning field to induce soft part segmentation. 
To support category-level generalization, our \OURS introduces a shared canonical template with instance-specific residual shape that decouples geometry from articulation across objects with diverse shapes. 
We learn this canonical representation through cycle reconstruction between observed and canonical shapes and through cross-space alignment between each input and its canonical template. Additional regularizers are added on keypoints, joint directions and joint locations to guide the model towards physically meaningful articulation with minimal supervision.

\noindent We summarize our \textbf{main contributions} as follows:
\begin{itemize}
\item We introduce \OURS{}, a self-supervised framework that recovers canonical shape, rigid parts and joint parameters of articulated objects from a single 3D observation. 
\item \OURS{} disentangles shape from pose and models part motion in a unified blend-skinning deformation space. This design yields interpretable joint pivots, axes and articulation states. 
\item Experiments on synthetic and real-world benchmarks show that \OURS{} outperforms prior self-supervised methods in part segmentation, joint estimation and articulated pose accuracy.
\end{itemize}

\section{Related Work}
\label{sec:related_work}

\noindent \textbf{Modeling Articulated Objects.} 
Recent work has moved from modeling humans and animals~\cite{deng2020nasa,loper2023smpl,mihajlovic2021leap,noguchi2021neural,liu2023paris} to general articulated objects with rigid parts. Early methods such as A-SDF~\cite{mu2021sdf} and ANCSH~\cite{li2020category} use implicit fields to disentangle shape and articulation. CAPTRA~\cite{weng2021captra} and Ditto~\cite{jiang2022ditto} extend this line to dynamic tracking from point clouds. 
Multi-view approaches like PARIS~\cite{liu2023paris} and CLANeRF~\cite{tseng2022clanerf} reconstruct part-aware geometry but rely on dense 3D supervision. 
CARTO~\cite{heppert2023carto}, NASAM~\cite{wei2022self} and REACTO~\cite{song2024reacto} adopt weak or self-supervision to reduce annotation cost. However, they still assume stereo inputs or observable motion. 
DigitalTwinArt~\cite{weng2024DigitalTwinArt} uses paired frames with known articulation changes, and recent methods such as ArticulatedGS~\cite{guo2025articulatedgs} and IAAO~\cite{zhang2025iaao} exploit 3D Gaussian splats from multi-frame data. 
In contrast, our \OURS recovers joints, parts and canonical shapes from a single RGB-D or point cloud input without templates, annotations or motion sequences.
Consequently, our single-view label-free design supports more scalable deployment.

\smallskip
\noindent \textbf{Category-Level Articulated Object Pose Estimation.}
Prior work tackles this task with several strategies. PartMobility~\cite{shi2021PartMobility} and UPPD~\cite{kawana2022uppd} rely on multi-view inputs or shape annotations, which restricts them to settings beyond single-frame inference. Other methods~\cite{lei2022cadex,mu2021asdf,paschalidou2021neural} use implicit shape modeling or active exploration~\cite{hausman2015active,jiang2022ditto} and require dense labels or complex data collection. EAP~\cite{liu2023EAP} proposes a self-supervised pipeline from single views but has slow inference and unstable segmentation. The more recent work OP-Align~\cite{che2024opalign} improves alignment through joint object- and part-level canonicalization. However, it still struggles with large shape variation and fine-grained articulation. 
Our method advances this line by using pose-invariant features and a blend-skinning-based deformation model to infer part motion and canonical shape from a single view without annotations and with stronger generalization across category-level variation.

\section{Our Methodology}
\label{sec:method}

\noindent \textbf{Objective.}
Given an input point cloud $\mathbf{X} \in \mathbb{R}^{3 \times N}$ of an articulated object with $P$ rigid parts, we aim to segment points into part-level components, estimate the rigid pose of each part, and recover joint parameters that describe the relative motion between parts.  
We consider $J$ joints that connect pairs of parts and associate each joint $j$ with a pivot $\mathbf{c}^{[j]} \in \mathbb{R}^3$, a motion axis $\mathbf{d}^{[j]} \in \mathbb{R}^3$ and a pair of articulation states $\mathbf{a}^{[j]} \in \mathbb{R}^2$ with one entry per incident part.
Each part $p$ has a rigid transformation $\mathbf{T}^{[p]} = [\mathbf{R}^{[p]} \, | \, \mathbf{t}^{[p]}] \in \mathrm{SE}(3)$ that shares the appropriate joint pivot with its adjacent part and differs in articulation state.  
Objects appear under arbitrary global poses in the observation space, and thus we define a pose-normalized canonical space where globally aligned shapes support consistent part and joint analysis.

\begin{figure*}[t]
\centering
\includegraphics[scale=0.42]{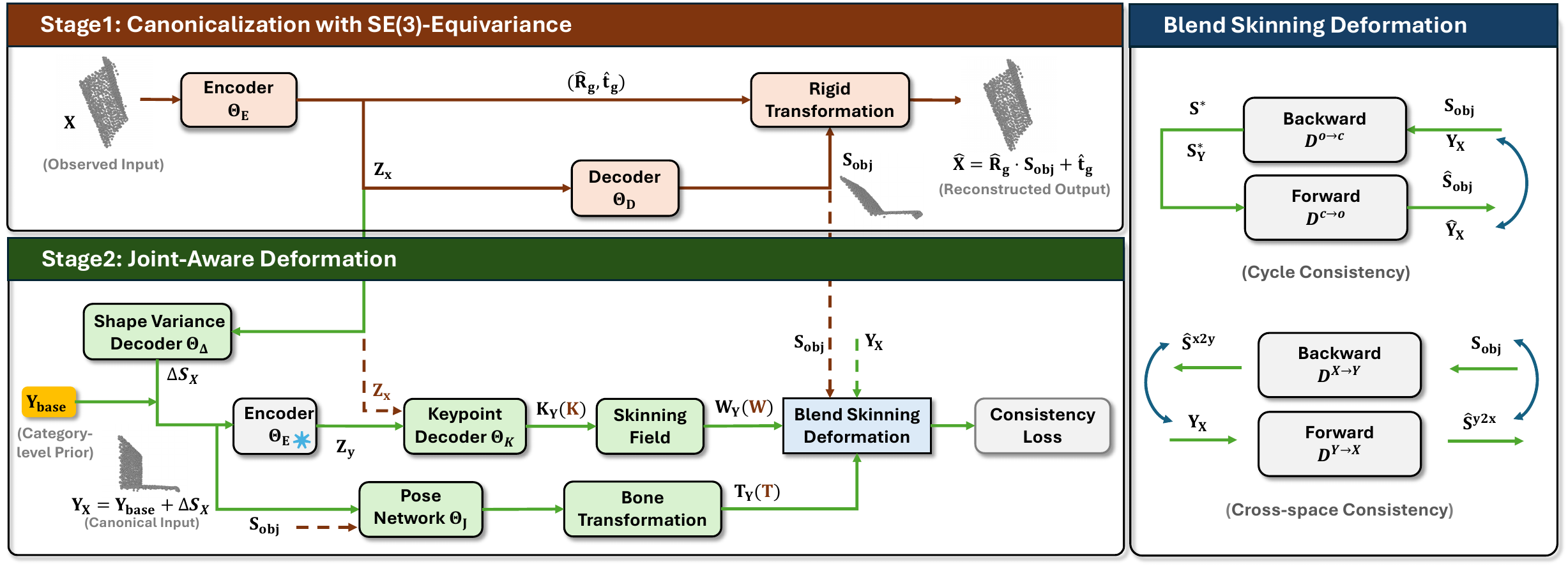}
\caption{ \textbf{Our SCAPO framework.} Stage 1 canonicalizes the input point cloud $\mathbf{X}$ using the global pose $\{\mathbf{R}_g, \mathbf{t}_g\}$ predicted by a SE(3)-equivariant auto-encoder ($\Theta_E$ and $\Theta_D$). In Stage 2, $\Theta_J$ takes as input the globally aligned shape $\mathbf{S}_{\text{obj}}$ and the learnable canonical template $\mathbf{Y}_X$ to estimate joint parameters and derive bone transformations. The affine-invariant features $\mathbf{Z}_x$ are fed to $\Theta_K$ and $\Theta_\Delta$ to learn keypoints $\mathbf{K}$ and shape variance $\Delta S_x$. With the skinning weights $\mathbf{W}$ and bone transformations $\mathcal{D}$, our blend skinning module reconstructs coherent articulated shapes across poses. 
} \vspace{-4mm}
\label{fig:framework}
\end{figure*}

\smallskip
\noindent \textbf{Overview.}
%
Fig.~\ref{fig:framework} presents an overview of our \OURS framework, which consists of three main stages: 
1) The \textbf{\textit{canonicalization module}} (\cf Sec.~\ref{sec:canonicalization}) maps each input object into a shared canonical space where objects are globally aligned. It uses a SE(3)-equivariant auto-encoder to estimate the global pose and reconstruct the canonical shape from pose-invariant features with a self-supervision objective.
2) The \textbf{\textit{joint-aware deformation}} stage (\cf Sec.~\ref{sec:articulationModeling}) defines a bi-directional deformation pipeline that transforms points between the observation and canonical space with a blend skinning formulation. A keypoint network predicts part centroids to define a skeletal structure from which skinning weights are derived to guide deformation of all points.
A pose network estimates joint pivots, motion axes and articulation states that drive rigid transformations of individual parts.  
3) The \textbf{\textit{optimization and regularization}} stage (\cf Sec.~\ref{sec:optimization}) jointly refines part geometry and articulation under self-supervision.  
Based on the deformation pipeline, we use a cycle-consistency loss between observed and reconstructed shapes, and a cross-space consistency loss between canonical and observation coordinates.

\subsection{Canonicalization with SE(3)-Equivariance}
\label{sec:canonicalization} 

Articulated objects often exhibit large pose variation and intra-class shape differences. With only a single observation per instance, explicit motion cues are absent and articulation remains ambiguous. We thus introduce a \textbf{\textit{canonicalization module}} that learns a category-level canonical space, where diverse instances align to a common frame. The module uses an $\operatorname{SE}(3)$-equivariant auto-encoder $(\Theta_E, \Theta_D)$ built on Vector Neuron Networks (VNNs)~\cite{deng2021vector} and Vector Neuron Translation (VNT)~\cite{katzir2022shape}. Given a point cloud $\mathbf{X} \in \mathbb{R}^{N \times 3}$, the encoder produces affine-invariant features $\mathbf{Z}_{x}$ for estimation of the global rotation $\mathbf{R}_{g}$ and translation $\mathbf{t}_{g}$. The decoder maps the latent code to a canonical shape $\mathbf{S}_{\text{obj}}$ that represents the object in the shared canonical frame. This global pose and pose-invariant shape code support downstream part segmentation and articulation modeling.

\smallskip
\noindent \textbf{SE(3)-equivariant Encoder-Decoder.} 
The encoder $\Theta_E$ takes input $\mathbf{X}$ and progressively factors out global translation and rotation. 
We first estimates a translation vector $\mathbf{t}_{g}$ with VNT layers and builds translation-invariant features. We then applies VNN layers to extract orientation-aware features and aggregates them to predict the rotation matrix $\mathbf{R}_{g}$. 
This pose disentanglement produces a pose-invariant latent shape code $\mathbf{Z}_x$. 
The decoder $\Theta_D$ maps $\mathbf{Z}_x$ to a canonical shape $\mathbf{S}_{\text{obj}} \in \mathbb{R}^{N \times 3}$ that is independent of any global transformation. 
The original input $\mathbf{X}$ can be recovered as:
$\mathbf{X} = \mathbf{S}_{\text{obj}} \cdot \mathbf{R}_g + \mathbf{1}_N \cdot \mathbf{t}_g^\top$,
where $\mathbf{1}_N \in \mathbb{R}^{N \times 1}$ broadcasts the translation to all points. 
This guarantees that shape features are invariant to $\operatorname{SE}(3)$ transformations, while pose parameters remain equivariant.

\smallskip
\noindent \textbf{Training Objectives for Canonical Alignment.} 
We adopt the training objectives from~\cite{katzir2022shape} to achieve pose-shape disentanglement and stable canonical alignment. A reconstruction loss $\mathcal{L}_{\text{rec}}$ matches the input and the reconstructed output. An orthonormality loss $\mathcal{L}_{\text{ortho}}$ regularizes $\mathbf{R}_g$ towards $\mathrm{SO}(3)$. An augmentation consistency loss $\mathcal{L}_{\text{aug\_consist}}$ enforces pose consistency under input augmentations. A canonical consistency loss $\mathcal{L}_{\text{can\_consist}}$ promotes robustness to known $\operatorname{SE}(3)$ perturbations. We combine these terms as:
\begin{equation}\small
\mathcal{L} = \mathcal{L}_{\text{rec}} 
+ \lambda_1 \mathcal{L}_{\text{ortho}} 
+ \lambda_2 \mathcal{L}_{\text{aug\_consist}} 
+ \lambda_3 \mathcal{L}_{\text{can\_consist}},
\end{equation}
where $\lambda_1$, $\lambda_2$ and $\lambda_3$ are balancing coefficients. 
This stage aligns instances globally and supports part segmentation and articulation modeling.

\subsection{Joint-Aware Deformation}
\label{sec:articulationModeling}

We model articulation with a joint-aware blend-skinning framework in the canonical space. A keypoint network $\Theta_K$ predicts part centroids that anchor a set of bones and form a skeleton over the canonical shape. 
A skinning field then assigns soft weights that control the influence of each part on every 3D point. A pose network $\Theta_J$ predicts joint pivots, motion axes and articulation states. These quantities drive per-part rigid transforms ${\mathbf{T}^{[p]}}$ and a bidirectional blend-skinning map between observation and canonical spaces.

\smallskip
\noindent \textbf{Forward and Backward Deformation.} 
Given the canonical shape $\mathbf{S}_{\text{obj}}$, we assign a set of bones $\mathbf{B} = \{\mathbf{b}^{[p]}\}_{p=1}^P$, where each bone $\mathbf{b}^{[p]}$ corresponds to a locally rigid part such as a handle, door or lid. 
Each point $\mathbf{s}_i \in \mathbf{S}_{\text{obj}}$ receives influence from all bones with weights $\mathbf{w}_i \in \mathbb{R}^{P}$ from the learned skinning matrix $\mathbf{W} \in \mathbb{R}^{N \times P}$. 
For each bone $\mathbf{b}^{[p]}$, we define a rigid transformation $\mathbf{T}^{[p]} = [\mathbf{R}^{[p]} \, | \, \mathbf{t}^{[p]}] \in \mathrm{SE}(3)$ based on its joint parameters. 

We obtain blended per-point transformations in canonical-to-observation and observation-to-canonical directions as:
\begin{equation}
\mathbf{T}^{c \rightarrow o}_i = \sum_{p=1}^{P} \mathbf{w}^{[p]}_i \, \mathbf{T}^{[p]},
\quad
\mathbf{T}^{o \rightarrow c}_i = \sum_{p=1}^{P} \mathbf{w}^{[p]}_i \, (\mathbf{T}^{[p]})^{-1}.
\label{eq:bleneded_transform}
\end{equation}
The \textbf{\textit{backward deformation}} $\mathcal{D}^{o \rightarrow c}(\cdot)$ maps an observed point $\mathbf{s}_i$ to its canonical counterpart $\mathbf{s}^*_i$:
\begin{equation}
\mathbf{s}^*_i = \mathcal{D}^{o \rightarrow c}(\mathbf{s}_i)
= \mathbf{T}^{o \rightarrow c}_i
\begin{bmatrix} \mathbf{s}_i^\top & 1 \end{bmatrix}^\top,
\label{eq:inverse_deform}
\end{equation}
and the \textbf{\textit{forward deformation}} $\mathcal{D}^{c \rightarrow o}(\cdot)$ maps a canonical reference point $\mathbf{s}^*_i$ back to the observation:
\begin{equation}
\mathbf{s}_i = \mathcal{D}^{c \rightarrow o}(\mathbf{s}^*_i)
= \mathbf{T}^{c \rightarrow o}_i
\begin{bmatrix} \mathbf{s}_i^{* \top} & 1 \end{bmatrix}^\top.
\label{eq:forward_deform}
\end{equation}
This forward-backward pipeline supports cycle-consistent reconstruction. 
Each point can move between the observed shape and the canonical frame in both directions, and we use this property during training to enforce geometric consistency and coherent part motions.

\smallskip
\noindent \textbf{Bone Transformation from Joint Parameters.} 
We use a pose estimation network $\Theta_J$ to predict joint parameters for each part $p$. 
These parameters consist of the pivot point $\mathbf{c}^{[p]}$, motion axis $\mathbf{d}^{[p]}$ and scalar articulation state $a^{[p]}$. 
From these quantities, we construct the bone transformation $\mathbf{T}^{[p]}$ that models the motion of part $p$ relative to the canonical frame. 
For each part, the transformation applies in three steps: 1) translate points to the pivot $\mathbf{c}^{[p]}$, 2) apply the motion, and 3) translate them back. 
We define the rotation $\mathbf{R}^{[p]}$ and translation $\mathbf{t}^{[p]}$ as:
\begin{subequations} \small
\begin{align}
\mathbf{R}^{[p]} &=
\begin{cases}
\mathrm{ExpSO3}([\mathbf{d}^{[p]}]_\times \cdot a^{[p]}), & \text{if revolute}, \\
\mathbf{I}, & \text{if prismatic},
\end{cases} \\
\mathbf{t}^{[p]} &=
\begin{cases}
\mathbf{c}^{[p]}, & \text{if revolute}, \\
\mathbf{d}^{[p]} \cdot a^{[p]}, & \text{if prismatic}.
\end{cases}
\end{align}
\end{subequations}
where $[\cdot]_\times$ denotes the skew-symmetric matrix of a vector and $\mathrm{ExpSO3}(\cdot)$ maps an axis-angle vector to a rotation in $\mathrm{SO}(3)$. Combined with blend skinning, these transformations yield consistent interpretable deformations and preserve geometric fidelity across articulation states.

\smallskip
\noindent \textbf{Bone Anchor Discovery \& Skinning Field.} 
Following~\cite{yang2022banmo,song2024reacto}, we represent each bone $\mathbf{b}^{[p]}$ as a Gaussian with center $\mathbf{O}^{[p]} = \mathbf{k}^{[p]}$, orientation matrix $\mathbf{V}^{[p]} \in \mathbb{R}^{3 \times 3}$ and diagonal scale matrix $\boldsymbol{\Lambda}^{[p]} \in \mathbb{R}^{3 \times 3}$. 
The centers are from predicted keypoints $\mathbf{K} = \{\mathbf{k}^{[1]}, \ldots, \mathbf{k}^{[P]}\} \in \mathbb{R}^{P \times 3}$, which we learn as geometric part centroids. A lightweight MLP $\Theta_K$ predicts these keypoints from affine-invariant features $\mathbf{Z}_x$, and they anchor part-level deformations.

To model part influence, we define a skinning field $\boldsymbol{\Omega}: \mathbf{s}_i \mapsto \mathbf{w}_i \in \mathbb{R}^P$ that assigns soft bone weights to each point $\mathbf{s}_i$. 
We compute the unnormalized influence of bone $p$ as:
\begin{equation}\small
\mathbf{W}^{[p]}_i = (\mathbf{s}_i - \mathbf{O}^{[p]})^\top \mathbf{Q}^{[p]} (\mathbf{s}_i - \mathbf{O}^{[p]}), 
\end{equation}
where $\mathbf{Q}^{[p]} = \mathbf{V}^{[p]\top} \boldsymbol{\Lambda}^{[p]} \mathbf{V}^{[p]}$. 
This Mahalanobis distance measures how well a point fits the shape and orientation of bone $p$. 
We then obtain the final skinning weights with a temperature-controlled softmax:
$\mathbf{w}_i = \mathrm{softmax}\left( \frac{-\mathbf{W}_i}{\gamma} \right)$,
where $\gamma$ controls the sharpness of the assignment. 
Lower $\gamma$ yields sparse and localized associations so that each point mainly depends on nearby bones.

\smallskip
\noindent \textbf{Canonical Reconstruction via Cycle Consistency.} 
To obtain a consistent reference for each input $\mathbf{X}$, we first invert the predicted global pose and transform $\mathbf{X}$ into the aligned shape:
$\mathbf{S}_{\text{obj}} = \mathcal{T}^{-1}_{\text{global}}(\mathbf{X})$.
We then map $\mathbf{S}_{\text{obj}}$ to a canonical shape $\mathbf{S}^*$ that encodes consistent articulation states across instances and acts as the articulation anchor. We define the bidirectional deformation pipeline as:
\begin{align}\small
\mathbf{X} &\xrightarrow{\mathcal{T}^{-1}_{\text{global}}} \mathbf{S}_{\text{obj}} 
\xrightarrow{\mathcal{D}^{o \rightarrow c}} \mathbf{S}^* 
\xrightarrow{\mathcal{D}^{c \rightarrow o}} \hat{\mathbf{S}}_{\text{obj}} 
\xrightarrow{\mathcal{T}_{\text{global}}} \hat{\mathbf{X}}.
\end{align}
We then learn the canonical space through cycle reconstruction and cross-space alignment between the observations and canonical templates.

\smallskip
\noindent \textbf{1) \textit{Cycle Reconstruction.}}
To keep articulation and pose transformations invertible and consistent, we apply the forward deformation $\mathcal{D}^{c \rightarrow o}$ followed by the global transform $\mathcal{T}_{\text{global}}$ and reconstruct the input. 
We define the cycle consistency loss as: $\mathcal{L}_{\text{cycle}} = \| \hat{\mathbf{X}} - \mathbf{X} \|_1$.

We further regularize the reference $\mathbf{S}^*$ through a canonical template $\mathbf{Y}_{X}$ for each input $X$. 
This template captures both instance-specific geometry and a fully canonical pose:
$\mathbf{Y}_X = \Delta S_X + \mathbf{Y}_{\text{base}}$,
where $\Delta S_X = \Theta_\Delta(\mathbf{F}_x)$ denotes the residual shape of $X$ predicted from the decoder final feature map $\mathbf{F}_x$, and $\mathbf{Y}_{\text{base}}$ is a learnable category-level prior that encodes shared canonical geometry and pose. 
We apply the same cycle to $\mathbf{Y}_X$, \ie:
$\mathbf{Y}_X
\xrightarrow{\mathcal{D}^{o \rightarrow c}_{[Y]}} \mathbf{S}^{*}_Y 
\xrightarrow{\mathcal{D}^{c \rightarrow o}_{[Y]}} \hat{\mathbf{Y}}_X$,
and supervise it with $\mathcal{L}_{\text{cycle}}$ to keep deformations coherent between object and canonical spaces.

\smallskip
\noindent \textbf{2) \textit{Cross-space Alignment.}}
Beyond the bone transformations $\mathbf{T}^{[p]}$ used for within-instance cycles, we define cross-space bone transformations $\mathbf{T}^{[p]}_{\text{X}\rightarrow\text{Y}}$ and $\mathbf{T}^{[p]}_{\text{Y}\rightarrow\text{X}}$ to align the observed input $\mathbf{X}$ and its canonical template $\mathbf{Y}_X$. 
We use these maps in the deformation pipelines:
$\mathbf{Y}_X
\xrightarrow{\mathcal{D}^{\text{Y}\rightarrow\text{X}}} \hat{\mathbf{S}}^{y2x}, 
\mathbf{S}_{\text{obj}}
\xrightarrow{\mathcal{D}^{\text{X}\rightarrow\text{Y}}} \hat{\mathbf{S}}^{x2y}$.

Given joint parameters $(\mathbf{c}_x^{[p]}, \mathbf{d}_x^{[p]}, a_x^{[p]})$ for $\mathbf{S}_{\text{obj}}$ and $(\mathbf{c}_y^{[p]}, \mathbf{d}_y^{[p]}, a_y^{[p]})$ for $\mathbf{Y}_X$, we define the transformation $\mathbf{T}^{[p]}_{\text{X}\rightarrow\text{Y}}$ from input space to canonical space as:
\begin{subequations}
\small
\begin{align}
&\mathbf{R}^{[p]}_{\text{X}\rightarrow\text{Y}} = \mathbf{R}_{x2y}^{[p]} \cdot \mathbf{R}_{r,x2y}^{[p]}~~\text{with} \quad 
\mathbf{R}_{r,x2y}^{[p]} \cdot \mathbf{d}_x^{[p]} = \mathbf{d}_y^{[p]}, \\
&\mathbf{t}^{[p]}_{\text{X}\rightarrow\text{Y}} = \mathbf{t}_{x2y}^{[p]} + \mathbf{t}_{r,x2y}^{[p]}~~~~~\text{with} \quad 
\mathbf{t}_{r,x2y}^{[p]} = \mathbf{c}_y^{[p]}.
\end{align}
\end{subequations}
and
\begin{subequations}\small
\begin{align}
\mathbf{R}_{x2y}^{[p]} &=
\begin{cases}
\mathrm{ExpSO3}([\mathbf{d}_y^{[p]}]_\times \cdot (a_y^{[p]} - a_x^{[p]})) & \text{if revolute}, \\
\mathbf{I} & \text{if prismatic},
\end{cases} \\
\mathbf{t}_{x2y}^{[p]} &=
\begin{cases}
\mathbf{0} & \text{if revolute}, \\
\mathbf{d}_y^{[p]} \cdot (a_y^{[p]} - a_x^{[p]}) & \text{if prismatic}.
\end{cases}
\end{align}
\end{subequations}
Here, $\mathbf{R}_{r,x2y}^{[p]}$ aligns the input joint axis $\mathbf{d}_x^{[p]}$ with the canonical axis $\mathbf{d}_y^{[p]}$, and $\mathbf{R}_{x2y}^{[p]}$ encodes the relative motion between articulation states.  
We construct $\mathbf{T}^{[p]}_{\text{Y}\rightarrow\text{X}}$ analogously.  
These cross-space transformations define the blended transforms in $\mathcal{D}^{\text{Y}\rightarrow\text{X}}$ and $\mathcal{D}^{\text{X}\rightarrow\text{Y}}$ via Eq.~\ref{eq:bleneded_transform}.

We further align the deformed shapes with the learned reconstructions using a shape reconstruction loss:
\begin{equation}\small
\mathcal{L}_{\text{recon}} = \operatorname{CD}(\hat{\mathbf{S}}^{x2y}, \mathbf{Y}_X), 
\quad 
\mathcal{L}_{\text{recon}}^Y = \operatorname{CD}(\hat{\mathbf{S}}^{y2x}, \mathbf{S}_{\text{obj}}),
\end{equation}
where $\operatorname{CD}(\cdot)$ is the Chamfer distance. 
$\mathcal{L}_{\text{recon}}$ aligns canonicalized input shape with canonical template $\mathbf{Y}_X$, and $\mathcal{L}_{\text{recon}}^Y$ enforces that $\mathbf{Y}_X$ can deform back to aligned observation. 

\medskip
\noindent The combined losses $\mathcal{L}_{\text{cycle}}$, $\mathcal{L}_{\text{recon}}$ and $\mathcal{L}_{\text{recon}}^Y$ encourage the canonical template $\mathbf{Y}_X$ to encode well-aligned geometry and guide the joint parameters towards accurate reversible articulation-aware deformations.

\subsection{Optimization and Regularization}
\label{sec:optimization}

This stage jointly refines part geometry and articulation parameters under self-supervision. We train the full framework end-to-end and optimize global pose, articulation parameters, skinning weights and canonical shapes with the reconstruction objectives described above. 
To obtain a meaningful decomposition of shape and articulation and to stabilize training, we add several regularization losses that reflect the structure of articulated objects.

\smallskip
\noindent\textbf{Keypoint Segmentation Consistency.}
We encourage each predicted keypoint (bone position) to match the geometric center of its associated part. Given the canonicalized shape $\mathbf{S}_{\text{obj}} = \{\mathbf{s}_i\}_{i=1}^N \in \mathbb{R}^{N \times 3}$ and the part segmentation matrix (skinning weights) $\mathbf{W} = \{w_i^{[p]}\} \in \mathbb{R}^{N \times P}$, we define the soft centroid of part $p$ as:
\begin{equation}\small
\mathbf{m}^{[p]} = \frac{1}{N_p} \sum_{i=1}^{N} w_i^{[p]} \cdot \mathbf{s}_i,
\quad
N_p = \sum_{i=1}^N w_i^{[p]}.
\end{equation}
We then penalize the deviation between each keypoint $\mathbf{k}^{[p]} \in \mathbf{K}$ and its part centroid:
\begin{equation}\small
\mathcal{L}_{\text{kp-seg}} = \frac{1}{P} \sum_{p=1}^{P} 
\mathbb{I}(N_p > \tau) \cdot 
\| \mathbf{k}^{[p]} - \mathbf{m}^{[p]} \|_2^2,
\end{equation}
where $\tau$ is a minimum support threshold that avoids unstable gradients from under-supported parts. 
We apply this loss on $\mathbf{S}_{\text{obj}}$, $\mathbf{Y}_X$ and $\mathbf{S}^*$.

\smallskip
\noindent\textbf{Keypoint-Supervised Segmentation Loss.}
We further regularize the part segmentation with nearest-keypoint supervision. 
For each point $\mathbf{s}_i$ in $\mathbf{S}_{\text{obj}}$, we find its closest keypoint:
\begin{equation}\small
p^*_i = \arg\min_{p} \|\mathbf{s}_i - \mathbf{k}^{[p]}\|_2^2,
\end{equation}
convert this index into a one-hot vector $\mathbf{y}_i \in \{0,1\}^P$, and define:
\begin{equation}\small
\mathcal{L}_{\text{seg}} = \frac{1}{N} \sum_{i=1}^N \| \mathbf{w}_i - \mathbf{y}_i \|_2^2.
\end{equation}
This objective encourages each point to take high weight on the part whose keypoint lies nearest to it and provides geometric supervision even without dense labels. 
We also use the same $\mathbf{y}_i$ to supervise the weights $\mathbf{w}_i^*$ on the fully canonicalized shape $\mathbf{S}^*$.

\smallskip
\noindent\textbf{Canonical Shape and Joint Regularization.}
We regularize the canonical shape and joint parameters with three additional terms:

\smallskip
\noindent \textbf{1) \textit{Shape Variance.}}
We constrain the instance-specific residual shape $\Theta_\Delta(\cdot)$ to avoid excessive deformation. 
Let $\mathbf{V} \in \mathbb{R}^{B \times N \times 3}$ be the predicted residuals. 
We define
\begin{equation}\small
\mathcal{L}_{\text{shape-var}} = 1 - \exp\left(-\lambda_{\text{var}} \cdot \| \mathbf{V} \|_2^2 \right),
\end{equation}
with scaling factor $\lambda_{\text{var}}$ (e.g., 60). 
This loss keeps residuals compact but still allows fine-grained instance variation.

\smallskip
\noindent \textbf{2) \textit{Joint Direction.}}
We encourage each joint axis $\mathbf{d}^{[j]} \in \mathbb{R}^3$ to align with the dominant geometric variation near part boundaries. 
For each sample, we detect boundary points $\mathbf{s}_i$ with high segmentation entropy and compute the principal direction $\tilde{\mathbf{d}}$ via PCA on these points. 
We then define:
\begin{equation}\footnotesize
\mathcal{L}_{\text{dir-align}} = \frac{1}{B} \sum_{b=1}^B \left(1 - \left| \left\langle \mathbf{d}^{[j]}, \tilde{\mathbf{d}} \right\rangle \right| \right),
\end{equation}
where $\langle \cdot, \cdot \rangle$ denotes cosine similarity. 
The absolute value handles sign ambiguity and enforces symmetry-aware alignment. 
We apply this loss to joint directions on $\mathbf{S}^*$ and on the reference template $\mathbf{Y}_X$.

\smallskip
\noindent \textbf{3a) \textit{Joint-to-Shape Proximity.}}
We prevent joint pivots from drifting far from the surface by constraining their distance to the canonical shape. 
Let $\mathbf{C} = \{\mathbf{c}^{[j]}\}$ denote the pivots, and we define:
\begin{equation}\footnotesize
\mathcal{L}_{\text{joint-prox}} = 1 - \exp\left(-\lambda_{\text{prox}} \cdot \mathbb{E}_{p}\left[\min_{\mathbf{s}_i \in \mathbf{S}^*} \| \mathbf{c}^{[j]} - \mathbf{s}_i \|^2 \right] \right),
\end{equation}
with scaling factor $\lambda_{\text{prox}}$, \eg 30. 
We approximate the inner minimum with $k$-NN search and apply the same loss to pivots predicted on $\mathbf{Y}_X$.

\smallskip
\noindent \textbf{3b) \textit{Joint-to-Boundary Attraction.}}
We encourage joint pivots to lie near articulation boundaries. 
We use entropy as a boundary cue since regions close to part interfaces tend to have high segmentation uncertainty.  
For each point, we compute:
\begin{equation}\small
H_i = - \sum_{p=1^P} w_i^{[p]} \log w_i^{[p]}.
\end{equation}
We then define a soft attention from each joint pivot $\mathbf{c}^{[j]}$ to nearby points ${\{\mathbf{s}_i\}}_{i=1}^N$:
\begin{equation}\small
\alpha_{j,i} = \frac{\exp(-\| \mathbf{c}^{[j]} - \mathbf{s}_i \|)}{\sum_k \exp(-\| \mathbf{c}^{[j]} - \mathbf{s}_k \|)},
\end{equation}
and finally, the joint–boundary loss is given as:
\begin{equation}\small
\mathcal{L}_{\text{joint-boundary}} = - \frac{1}{J} \sum_{j=1}^J \sum_{i=1}^N \alpha_{j,i} \cdot H_i.
\end{equation}
This term attracts pivots towards regions with high boundary entropy and thus towards articulation points. 
We apply the same regularization to pivots defined on the reference frame $\mathbf{Y}_X$.

\section{Experiment}
\label{sec:experiment}


\noindent \textbf{Datasets.}
We evaluate \OURS on synthetic and real-world datasets for canonical shape reconstruction, part segmentation, and joint parameter estimation under diverse articulation states. The \textbf{\textit{synthetic dataset}} follows EAP~\cite{liu2023EAP} and uses selected categories from HOI4D~\cite{liu2022hoi4d} and Shape2Motion~\cite{wang2019shape2motion}. It covers five articulated categories: \{\textit{laptop}, \textit{safe}, \textit{oven}, \textit{washer}, \textit{eyeglasses}\}. Each mesh is rendered into partial point clouds that mimic single-view depth observations, which introduces realistic occlusion and view-dependent visibility effects. The \textbf{\textit{real-world dataset}} from~\cite{che2024opalign} contains RGB-D scans of five categories without mesh ground truth: \{\textit{basket}, \textit{laptop}, \textit{suitcase}, \textit{drawer}, and \textit{scissors}\}. 3D point clouds are reconstructed by fusing depth with instance masks predicted by detection models. This setup provides a practical test of robustness under noisy real-world conditions.

\medskip
\noindent \textbf{Evaluation Metrics.} 
We follow prior work~\cite{li2020category,liu2023EAP,che2024opalign} and evaluate our \OURS along three dimensions: 1) \textbf{\textit{Part-level metrics:}} we report rotation error (degrees) and translation error of each part SE(3) pose, and the 3D Intersection-over-Union (IoU) of part segmentation masks. 2) \textbf{\textit{Articulation-state metrics:}} for revolute joints, we compute joint angle error (degrees); for prismatic joints, we measure translation magnitude error. 3) \textbf{\textit{Joint-parameter metrics:}} we evaluate axis orientation error (angular deviation) and pivot localization error (line-to-line distance) to capture both direction and position. For real-world data, we adopt \textbf{\textit{multi-threshold mean Average Precision}} (mAP). A part pose is correct if its rotation error is below $5^\circ$, $10^\circ$, or $15^\circ$ and its translation error is below 5, 10, or 15\,cm. We apply the same thresholds to joint pivot and axis estimation. 
For segmentation, we report mean IoU over parts under 75\% and 50\% thresholds.

\medskip
\noindent \textbf{Baselines.} 
For the synthetic dataset, we compare \OURS{} with three main methods: 
A self-supervised approach for part-level articulation EAP~\cite{liu2023EAP}; 
A supervised baseline for category-level 6D pose 3DGCN~\cite{lin2020convolution}; 
A recent self-supervised method for articulated object pose without pose or shape annotations OP-Align~\cite{che2024opalign}. 
We also include ICP and NPCS~\cite{li2020category} with EPN~\cite{chen2021equivariant} as backbone to cover classical alignment and joint estimation pipelines.

\medskip
\noindent \textbf{Implementation Details.}
We uniformly sample $N = 1024$ points per object for all experiments. 
Stage~1 trains the $\operatorname{SE}(3)$-equivariant auto-encoder with balancing weights $\lambda_1{=}\lambda_2{=}\lambda_3{=}0.1$. 
Stage~2 optimizes the articulation-aware modules: a PointNet-style joint parameter predictor, a keypoint detector, a category-level shape variance module, and other learnable priors. They are all implemented as MLP-based heads. We set the loss weights as 
$\lambda_{\text{cycle}}{=}10$, 
$\lambda_{\text{recon}}{=}10$, 
$\lambda_{\text{kp-seg}}{=}1$, 
$\lambda_{\text{seg}}{=}1$, 
$\lambda_{\text{shape-var}}{=}10$, 
$\lambda_{\text{dir-align}}{=}0.1$, 
$\lambda_{\text{joint-prox}}{=}1$, 
$\lambda_{\text{joint-boundary}}{=}3$. 
Optimization uses Adam with weight decay $1\times10^{-8}$ and cosine annealing over 200 epochs. 
The learning rate is $1\times10^{-3}$ in Stage~1 and $1\times10^{-4}$ in Stage~2, with a minimum of $1\times10^{-6}$. 
All models train on a single NVIDIA RTX~A5000 GPU with PyTorch.

\begin{table*}[t]
\centering
\scriptsize
\setlength{\tabcolsep}{2pt}
\renewcommand{\arraystretch}{0.6}
\caption{Results on synthetic datasets HOI4D and Shape2Motion. 
\textbf{S}, \textbf{D}, \textbf{C}, \textbf{R}, \textbf{t} denote part segmentation IoU; joint direction error (degree); joint pivot error; mean part rotation error (degree); and mean part translation error.} \vspace{-2mm}
\label{tab:synthetic_results}
\begin{tabular*}{\textwidth}{@{\extracolsep{\fill}}llcccrrrrrr}
\toprule
Metric & Method & \multicolumn{3}{c}{Supervision} & \multicolumn{2}{c}{HOI4D} & \multicolumn{3}{c}{Shape2Motion} & Mean \\
& & Pose & Segmentation & Joint & Laptop & Safe & Eyeglasses & Oven & Washer & \\
\midrule
\multirow{5}{*}{\textbf{S $\uparrow$}} 
& 3DGCN    & \checkmark & \checkmark & \checkmark & 99.28 & 94.11 & 81.93 & 94.46 & 93.15 & 92.59 \\
& ICP      &       & \checkmark &            & 59.96 & 66.90 & 49.49 & 75.17 & 72.80 & 64.46 \\
& EAP      &    &      &         & 86.04 & 44.64 & 62.84 & 76.22 & 73.27 & 68.20 \\
& OP-Align &            &            &            & 92.38 & 86.14 & 66.60 & 90.84 & 75.25 & 82.24 \\
& Ours     &            &            &            & 93.41 & 89.83 & 68.19 & 93.10 & 78.35 & 84.98 \\
\midrule
\multirow{5}{*}{\textbf{D $\downarrow$}} 
& 3DGCN    & \checkmark & \checkmark & \checkmark &  1.73 &  4.79 & 14.01 &  5.85 &  9.85 &  7.65 \\
& NPCS-EPN & \checkmark & \checkmark & \checkmark & 12.25 & 11.23 &  7.42 &  5.04 &  5.66 &  8.12 \\
& EAP      &            &            &            & 18.02 & 55.16 & 17.75 & 20.30 & 28.40 & 27.13 \\
& OP-Align &            &            &            &  1.46 &  1.34 &  3.28 &  4.41 & 10.47 &  4.60 \\
& Ours     &            &            &            &  1.41 &  1.22 &  3.14 &  4.70 &  8.97 &  3.89 \\
\midrule
\multirow{5}{*}{\textbf{C $\downarrow$}} 
& 3DGCN    & \checkmark & \checkmark & \checkmark & 0.014 & 0.024 & 0.096 & 0.071 & 0.107 & 0.062 \\
& NPCS-EPN & \checkmark & \checkmark & \checkmark & 0.134 & 0.084 & 0.096 & 0.076 & 0.078 & 0.094 \\
& EAP      &            &            &            & 0.170 & 0.170 & 0.087 & 0.090 & 0.118 & 0.127 \\
& OP-Align &            &            &            & 0.080 & 0.066 & 0.049 & 0.092 & 0.231 & 0.104 \\
& Ours     &            &            &            & 0.076 & 0.052 & 0.045 & 0.086 & 0.114 & 0.075 \\
\midrule
\multirow{5}{*}{\textbf{R $\downarrow$}} 
& 3DGCN    & \checkmark & \checkmark & \checkmark &  2.90 &  5.50 & 17.63 & 15.71 & 14.03 & 11.55 \\
& NPCS-EPN & \checkmark & \checkmark & \checkmark &  8.07 &  6.04 &  6.34 &  6.41 &  5.71 &  6.51 \\
& ICP      &            & \checkmark &            & 36.42 & 45.50 & 34.90 & 46.79 & 53.75 & 43.87 \\
& EAP      &            &            &            &  7.71 & 18.65 & 10.99 &  5.91 & 13.38 & 11.73 \\
& OP-Align &            &            &            &  2.87 &  6.54 &  4.65 &  5.76 & 11.83 &  6.73 \\
& Ours     &            &            &            &  2.17 &  6.26 &  4.95 &  5.60 &  9.93 &  5.78 \\
\midrule
\multirow{5}{*}{\textbf{t $\downarrow$}} 
& 3DGCN    & \checkmark & \checkmark & \checkmark & 0.018 & 0.025 & 0.177 & 0.112 & 0.122 & 0.091 \\
& NPCS-EPN & \checkmark & \checkmark & \checkmark & 0.048 & 0.028 & 0.066 & 0.028 & 0.194 & 0.073 \\
& ICP      &            & \checkmark &            & 0.293 & 0.264 & 0.155 & 0.081 & 0.079 & 0.174 \\
& EAP      &            &            &            & 0.079 & 0.065 & 0.070 & 0.061 & 0.062 & 0.067 \\
& OP-Align &            &            &            & 0.092 & 0.064 & 0.184 & 0.123 & 0.110 & 0.115 \\
& Ours     &            &            &            & 0.082 & 0.057 & 0.128 & 0.079 & 0.083 & 0.086 \\
\bottomrule
\end{tabular*}\vspace{-4mm}
\end{table*}

\subsection{Comparison with Baselines}

\noindent \textbf{Results on the Synthetic Dataset.} 
Table~\ref{tab:synthetic_results} reports results on HOI4D and Shape2Motion. 
Among \textbf{\textit{self-supervised}} ($<3\times$ $\checkmark$) methods, \OURS{} achieves the highest mean segmentation IoU (84.98\%) and the lowest errors on joint direction (3.89), joint pivot (0.075), and part rotation (5.78), with consistent improvements over OP-Align (82.24, 4.60, 0.104, 6.73). For translation, our error (0.086) is lower than OP-Align (0.115) and ICP (0.174), and close to the best self-supervised result from EAP (0.067). 
Gains are especially clear on \textit{eyeglasses} and \textit{washer}, where our \OURS{} improves both segmentation IoU and articulation errors over OP-Align. Compared with \textbf{\textit{fully supervised}} ($3\times$ $\checkmark$) 3DGCN and NPCS-EPN, our \OURS{} attains substantially lower joint direction and rotation errors, and competitive pivot and translation errors despite using no annotations.

\medskip
\noindent \textbf{Results on the Real-World Dataset.} 
Table~\ref{tab:real_world_results} reports results on the real-world benchmark under the same protocol as OP-Align. Across \textbf{\textit{all categories}}, \OURS{} surpasses OP-Align (SJ) in mean segmentation mIoU (91.89\%/98.29\% vs.\ 90.02\%/97.18\%) and mean joint mAP (62.51/89.59/95.91 vs.\ 61.37/87.72/92.73 at 5$^\circ$5cm–15$^\circ$15cm). It also attains higher part mAP at 5$^\circ$5cm and 15$^\circ$15cm, with comparable performance at 10$^\circ$10cm. 
These gains show that our self-supervised articulation model transfers well from synthetic to real data. 
Compared with \textbf{\textit{fully supervised}} (with Pose supervision) 3DGCN and PointNet++, our \OURS{} delivers higher mean joint mAP at all thresholds (62.51/89.59/95.91 vs.\ 47.51/85.79/94.59 for 3DGCN), and competitive or better part mAP at 5$^\circ$5cm and 15$^\circ$15cm. 
These results indicate that explicit pose supervision is not necessary once the model learns a robust canonical space and deformation-based alignment from data.

\begin{table*}[t]
\centering
\scriptsize
\setlength{\tabcolsep}{2pt}
\renewcommand{\arraystretch}{0.6}
\caption{Average precision results on the real-world dataset. \textit{Supervision} refers to the annotations used in training. \textbf{Segmentation}$^\uparrow$ denotes mIoU under 75\%/50\% thresholds. \textbf{Joint}$^\uparrow$ measures mAP under rotation/translation thresholds (5$^\circ$5cm–15$^\circ$15cm). \textbf{Part}$^\uparrow$ measures part-level mAP under the same thresholds.} \vspace{-1mm}
\label{tab:real_world_results}
\begin{tabular*}{\textwidth}{@{\extracolsep{\fill}}llccccccccccc}
\toprule
Category & Method & \multicolumn{3}{c}{Supervision} &
\multicolumn{2}{c}{Segmentation$^\uparrow$ (mIoU)} &
\multicolumn{3}{c}{Joint$^\uparrow$} &
\multicolumn{3}{c}{Part$^\uparrow$ (mAP)}  \\
& & Pose & Segmentation & Joint &
75\% & 50\% &
5$^\circ$5cm & 10$^\circ$10cm & 15$^\circ$15cm &
5$^\circ$5cm & 10$^\circ$10cm & 15$^\circ$15cm \\
\midrule
\multirow{4}{*}{\textbf{Basket}}
& 3DGCN       & \checkmark & \checkmark & \checkmark & 67.48 & 92.87 & 38.75 & 84.41 & 93.54 &  1.34 & 20.71 & 44.77 \\
& PointNet++  & \checkmark & \checkmark & \checkmark &  0.00 &  6.24 &  3.12 & 31.18 & 57.23 &  0.45 &  1.56 & 13.59 \\
& OP-Align(SJ)&            & \checkmark & \checkmark & 83.74 & 98.22 & 60.13 & 91.31 & 95.76 &  0.00 & 18.71 & 51.45 \\
& Ours        &            & \checkmark & \checkmark & 86.64 & 99.11 & 62.36 & 93.32 & 96.88 &  0.00 & 19.60 & 52.34 \\
\midrule
\multirow{4}{*}{\textbf{Drawer}}
& 3DGCN       & \checkmark & \checkmark & \checkmark & 82.52 & 96.90 & 38.27 & 76.32 & 91.15 & 28.00 & 68.36 & 88.50 \\
& PointNet++  & \checkmark & \checkmark & \checkmark &  3.22 & 40.46 & 16.09 & 42.53 & 64.37 &  8.74 & 42.76 & 65.98 \\
& OP-Align(SJ)&            & \checkmark & \checkmark & 88.74 & 95.17 & 54.71 & 82.76 & 85.52 & 17.93 & 52.18 & 85.86 \\
& Ours        &            & \checkmark & \checkmark & 91.03 & 96.09 & 56.55 & 85.29 & 87.36 & 19.31 & 54.71 & 88.28 \\
\midrule
\multirow{4}{*}{\textbf{Laptop}}
& 3DGCN       & \checkmark & \checkmark & \checkmark & 93.20 & 95.87 & 73.30 & 96.12 & 96.60 & 31.31 & 72.57 & 82.03 \\
& PointNet++  & \checkmark & \checkmark & \checkmark & 91.99 & 96.84 & 45.63 & 88.35 & 92.23 & 11.89 & 51.94 & 72.57 \\
& OP-Align(SJ)&            & \checkmark & \checkmark & 97.82 & 98.54 & 91.17 & 97.33 & 98.06 & 36.17 & 73.54 & 90.53 \\
& Ours        &            & \checkmark & \checkmark & 98.79 & 99.51 & 92.23 & 99.27 & 98.30 & 37.38 & 74.51 & 92.00 \\
\midrule
\multirow{4}{*}{\textbf{Scissors}}
& 3DGCN       & \checkmark & \checkmark & \checkmark & 76.01 & 94.54 & 43.94 & 85.99 & 97.15 &  1.66 & 22.33 & 50.83 \\
& PointNet++  & \checkmark & \checkmark & \checkmark &  0.00 & 17.58 & 19.71 & 57.72 & 83.13 &  0.24 &  2.14 &  5.23 \\
& OP-Align(SJ)&            &     \checkmark       & \checkmark & 94.77 & 99.21 & 53.21 & 97.39 & 99.38 &  9.54 & 45.65 & 64.55 \\
& Ours        &            & \checkmark & \checkmark & 95.49 & 99.76 & 53.92 & 98.57 & 99.52 & 10.69 & 47.27 & 66.03 \\
\midrule
\multirow{4}{*}{\textbf{Suitcase}}
& 3DGCN       & \checkmark & \checkmark & \checkmark & 97.36 & 98.45 & 43.31 & 86.09 & 94.49 & 10.24 & 49.87 & 77.17 \\
& PointNet++  & \checkmark & \checkmark & \checkmark &  3.94 & 49.87 & 20.73 & 66.14 & 80.84 &  1.05 & 17.84 & 41.73 \\
& OP-Align(SJ)&            & \checkmark & \checkmark & 85.04 & 94.75 & 44.62 & 69.89 & 96.26 &  1.26 & 26.12 & 51.71 \\
& Ours        &            & \checkmark & \checkmark & 87.50 & 97.00 & 47.50 & 71.50 & 97.50 &  2.50 & 27.50 & 54.00 \\
\midrule
\multirow{4}{*}{\textbf{Mean}}
& 3DGCN       & \checkmark & \checkmark & \checkmark & 83.31 & 95.83 & 47.51 & 85.79 & 94.59 & 13.07 & 46.77 & 68.66 \\
& PointNet++  & \checkmark & \checkmark & \checkmark & 19.83 & 42.20 & 21.06 & 57.38 & 75.64 &  4.47 & 23.75 & 39.82 \\
& OP-Align(SJ)&            & \checkmark & \checkmark & 90.02 & 97.18 & 61.37 & 87.72 & 92.73 & 12.98 & 43.24 & 62.97 \\
& Ours        &            & \checkmark & \checkmark & 91.89 & 98.29 & 62.51 & 89.59 & 95.91 & 13.98 & 44.72 & 70.53 \\
\bottomrule
\end{tabular*}\vspace{-1mm}
\end{table*}

\begin{figure*}[t]
\centering
\includegraphics[scale=0.25]{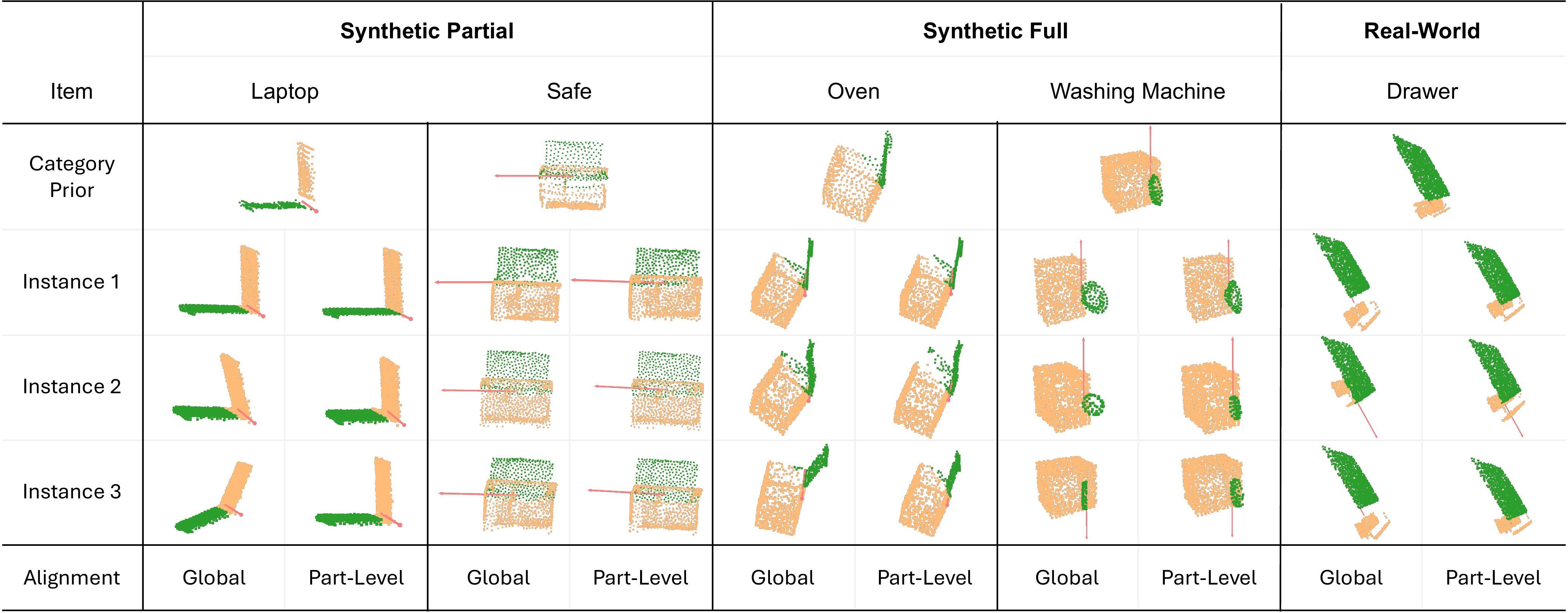}
\caption{
Qualitative results on synthetic partial (left), synthetic full (center), and real-world (right) point clouds. 
}
\label{fig:qualitative_part}
\vspace{-3mm}
\end{figure*}

\medskip
\noindent \textbf{Qualitative Results.} 
Fig.~\ref{fig:qualitative_part} illustrates some qualitative results of our \OURS{} on synthetic partial views, synthetic full observations, and real-world point clouds. On synthetic data, our model achieves precise object- and part-level alignment with clean segmentation and accurate joint prediction even under large pose changes. On real scans, we generalize well in the semi-supervised setting, align diverse inputs to a canonical frame, and recover plausible part motions. 
A key advantage of our \OURS{} is the tight coupling between segmentation and motion. 
Each part deforms consistently under its bone transform and joint parameters.
Segmentation noise therefore concentrates near boundaries, and interiors remain stable and semantically coherent.
Motion-guided optimization allows the network to exploit kinematic cues instead of depending purely on geometry. 
However, limitations appear when geometric evidence is weak, \eg under heavy occlusion or missing parts.
In these cases, the model may misestimate joints or exhibit segmentation drift.
These visual results support our claims on disentangling pose and shape, yielding consistent part segments, and reconstructing articulation with interpretable joints across categories.

\subsection{Ablation Study} 
We perform four ablations on the synthetic dataset to assess the impact of each regularization term in Tab.~\ref{tab:ablation}: 
a) Shape Variance Regularization $\mathcal{L}_{\text{shape-var}}$, 
b) Joint Direction Alignment $\mathcal{L}_{\text{dir-align}}$, 
c) Joint-to-shape Proximity $\mathcal{L}_{\text{joint-prox}}$, and 
d) Joint-to-boundary Attraction $\mathcal{L}_{\text{joint-boundary}}$. 
Removing $\mathcal{L}_{\text{shape-var}}$ hurts all metrics, with segmentation dropping from 84.98 to 80.15, joint direction error increasing from 3.89 to 10.63, and pivot error more than doubling from 0.075 to 0.175.  
Without $\mathcal{L}_{\text{dir-align}}$, the joint direction error rises to 8.54 and the pivot error to 0.086, which indicates less reliable articulation axes.  
Dropping $\mathcal{L}_{\text{joint-prox}}$ leads to a pivot error of 0.137 and a segmentation drop to 82.77, showing that pivots drift away from the actual geometry.  
Removing $\mathcal{L}_{\text{joint-boundary}}$ yields a pivot error of 0.113 with only mild changes in segmentation and direction error. 
These trends confirm that each loss term contributes to stable canonical shapes, accurate joint placement, and consistent part segmentation under self-supervised setting.

\begin{table}[t]
\centering
\caption{Ablation study on synthetic dataset.} \vspace{-3mm}
\begin{adjustbox}{width=0.47\textwidth}
\begin{tabular}{lccc}
\hline
\textbf{Method} & \textbf{Segmentation} $\uparrow$ & \textbf{Joint Direction Error} $\downarrow$ & \textbf{Joint Pivot Error $\downarrow$}  \\
\hline
Full       & 84.98 & 3.89 & 0.075   \\
w/o $\mathcal{L}_{\text{shape-var}}$ & 80.15 & 10.63. & 0.175   \\
w/o $\mathcal{L}_{\text{dir-align}}$ & 84.21 & 8.54 & 0.086  \\
w/o $\mathcal{L}_{\text{joint-prox}}$  & 82.77 & 4.73 & 0.137   \\
w/o $\mathcal{L}_{\text{joint-boundary}}$  & 84.50 & 4.12 & 0.113   \\
\hline
\end{tabular}
\end{adjustbox} 
\vspace{-6mm}
\label{tab:ablation}
\end{table}

\section{Conclusion}
\label{sec:conclusion}

We presented \OURS{}, a self-supervised framework for category-level articulated object modeling from a single 3D observation. 
It jointly estimates canonical shape, rigid parts and joint parameters. 
\OURS{} uses a learned canonical space with bone-based deformations to align instances, recover part motions and infer interpretable joints without manual labels or templates. 
Cycle and cross-space consistency losses help disentangle pose from shape and couple segmentation with motion. 
Parts remain stable in their interiors and vary mainly at boundaries. 
Regularization on shape variance, joint directions and joint locations stabilizes learning and improves geometric plausibility. 
Experiments on synthetic and real-world benchmarks show that \OURS{} delivers robust performance and clear articulation structure across diverse categories, often matching or surpassing supervised baselines. 
These results suggest that canonicalization with motion-aware deformation is an effective recipe for self-supervised articulated understanding.


\noindent\textbf{Limitations.}
Our method assumes a fixed number of joints and parts, which reduces flexibility across categories. The VNT auto-encoder can struggle under large shape variation or severe occlusion, which weakens canonical consistency. Joint inference also degrades with weak motion cues or noisy part segmentation.

\paragraph{Acknowledgment.} This research/project is supported by the Tier 2 grant MOET2EP20124-0015 from the Singapore Ministry of Education. 
{
    \small
    \bibliographystyle{ieeenat_fullname}
    \bibliography{main}
}


\end{document}